%% file: main.tex
\documentclass[letterpaper, 10 pt, conference]{ieeeconf}  

\IEEEoverridecommandlockouts                              

\makeatletter
\let\NAT@parse\undefined
\makeatother

\usepackage{graphics} 
\usepackage{epsfig} 
\usepackage{mathptmx} 
\usepackage{times} 
\usepackage{amsmath} 
\usepackage{amssymb}  

\usepackage{multirow}
\usepackage{array}
\usepackage{float}
\usepackage{dsfont}

\usepackage{booktabs}
\usepackage{algorithm}
\usepackage{algpseudocode}
\newcommand{\drule}{\specialrule{0.2pt}{1pt}{1pt} \specialrule{0.2pt}{0pt}{\belowrulesep}}
\usepackage{pifont}
\usepackage{xcolor}
\usepackage{pdfrender}
\usepackage{url}
\usepackage{breakurl}
\usepackage{afterpage}
\usepackage{mathtools}
\usepackage{hyperref}
\usepackage[capitalise]{cleveref} 
\usepackage{pifont}
\usepackage{balance}
\usepackage{microtype}

\usepackage[font=footnotesize]{caption}
\usepackage[labelformat=simple]{subcaption}

\usepackage{colortbl}

\usepackage[numbers,sort&compress]{natbib}

\hypersetup{
  bookmarksopen,
  bookmarksnumbered,
  pdfpagemode=UseOutlines,
  colorlinks=true,
  linkcolor=black,
  anchorcolor=blue,
  citecolor=black,
  filecolor=blue,
  menucolor=blue,
  urlcolor=blue
}

\title{\LARGE \bf
V-STRONG: Visual Self-Supervised Traversability Learning for Off-road Navigation
}
\author{Sanghun Jung, JoonHo Lee, Xiangyun Meng, Byron Boots, and Alexander Lambert\\
University of Washington
\thanks{\hspace{-1em}Distribution Statement A. Approved for Public Release,
Distribution Unlimited.
}}

\begin{document}

\maketitle
\thispagestyle{empty}
\pagestyle{empty}

\input{sections/abstract}
\vspace{-0.1cm}
\input{sections/introduction}
\vspace{-0.4cm}
\input{sections/related_work}
\vspace{-0.2cm}
\input{sections/method}
\vspace{-0.25cm}
\input{sections/experiments}
\vspace{-0.3cm}
\input{sections/conclusion}
\vspace{-0.2cm}
\input{sections/acknowledgement}

\balance
\bibliographystyle{IEEEtran}
\bibliography{references}

\end{document}

%% file: sections/abstract.tex
\begin{abstract}
Reliable estimation of terrain traversability is critical for the successful deployment of autonomous systems in wild, outdoor environments. Given the lack of large-scale annotated datasets for off-road navigation, strictly-supervised learning approaches remain limited in their generalization ability. To this end, we introduce a novel, image-based self-supervised learning method for traversability prediction, leveraging a state-of-the-art vision foundation model for improved out-of-distribution performance. Our method employs contrastive representation learning using both human driving data and instance-based segmentation masks during training. We show that this simple, yet effective, technique drastically outperforms recent methods in predicting traversability for both on- and off-trail driving scenarios. We compare our method with recent baselines on both a common benchmark as well as our own datasets, covering a diverse range of outdoor environments and varied terrain types. We also demonstrate the compatibility of resulting costmap predictions with a model-predictive controller. Finally, we evaluate our approach on zero- and few-shot tasks, demonstrating unprecedented performance for generalization to new environments. Videos and additional material can be found here: \url{https://sites.google.com/view/visual-traversability-learning}.
\end{abstract}

%% file: sections/introduction.tex
\section{INTRODUCTION}
Autonomous navigation in off-road environments requires an accurate understanding of the terrain, particularly in identifying traversable areas by the vehicle. However, contrary to on-road scenarios~\cite{ade20k, bdd100k, mapillary, cityscapes}, the notion of traversability in unstructured outdoor settings is much more ambiguous, \textit{a priori}. For instance, vehicle interaction with terrain features such as ground-vegetation, rocks, and debris is strongly correlated to their size, shape, and material appearance. As such, manual assignment of traversability labels to off-road perception data is non-trivial and prone to error~\cite{daejeon2023}.
Furthermore, the complexity of terrain characteristics and the sheer variety of environments make comprehensive data collection of all terrain types a daunting task.
Although efforts have been made towards generating annotated, off-road datasets~\cite{valada2017deep, maturana2018real, wigness2019rugd, rellis, terrainnet}, these are generally restricted to a particular set of geospatial locations and seasonal conditions. 

For these reasons, achieving reliable performance in traversability prediction via supervised learning remains a challenging problem.
Despite the necessity for labeled datasets, recent work on leveraging semantic segmentation has demonstrated the viability of supervised methods for terrain classification~\cite{maturana2018real, bevnet, terrainnet}, typically by generating a Bird's Eye View (BEV) semantic map of the scene with projection of semantic labels~\cite{maturana2018real} or by directly predicting in BEV~\cite{bevnet, terrainnet}. Still, the resulting class predictions in such cases must be mapped to a traversability metric or cost, which may require manual tuning and/or additional geometric feature information.

In order to address the challenges of off-road traversability prediction, \textit{self-supervised} learning has shown to be a promising alternative~\cite{karnan2023self, castro2023does, daejeon2023, hutter2023, xue2023contrastive, jpl2022, scherer2023, wellhausen2020safe, wellhausen2019should, dahlkamp2006self, zurn2020self}. Instead of relying on manually annotated data, many of these approaches use projected traces of vehicle or robot trajectories derived from human-piloted examples as positive traversability labels. Although practical to generate, such datasets are only equipped with positive labels, where much of the observed terrain remains unlabeled. 
As a result, self-supervised methods that train single-class predictors can be prone to overfitting~\cite{daejeon2023}.
Additionally, most of these methods were tested on a restricted class of environments, many of which predominantly consist of on-trail scenarios, which is categorically similar to on-road evaluations~\cite{castro2023does, daejeon2023}. 

\begin{figure}[t!]
  \begin{subfigure}[b]{\linewidth}
    \centering
    \includegraphics[width=\textwidth]{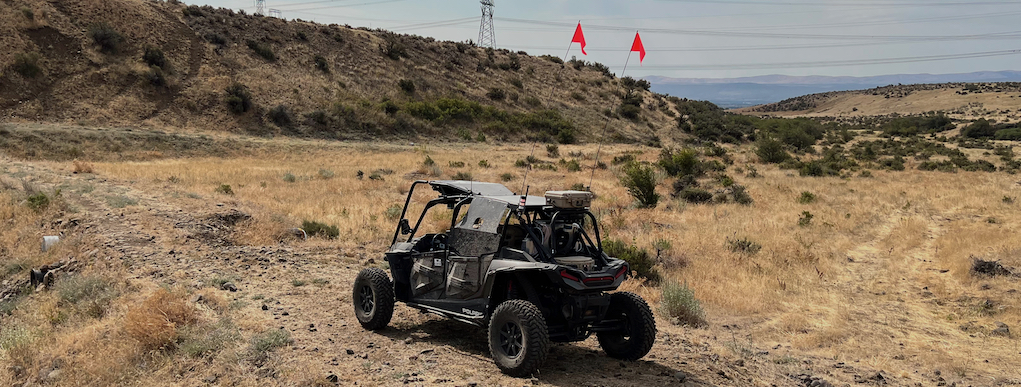}
    \vspace{-0.5\baselineskip}
    \label{vehicle}
  \end{subfigure} \\
   \begin{subfigure}[b]{0.325\linewidth}
    \centering
    \includegraphics[width=\textwidth]{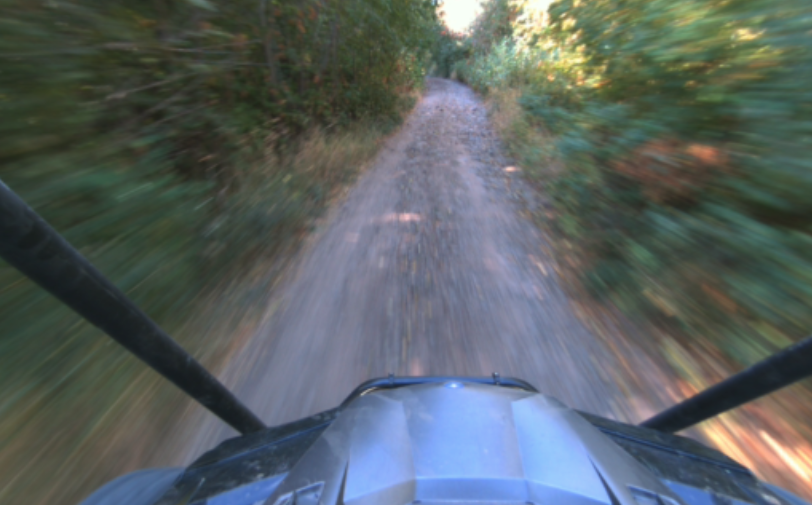}
    \label{subfig:rgb}
  \end{subfigure}
   \begin{subfigure}[b]{0.325\linewidth}
    \centering
    \includegraphics[width=\textwidth]{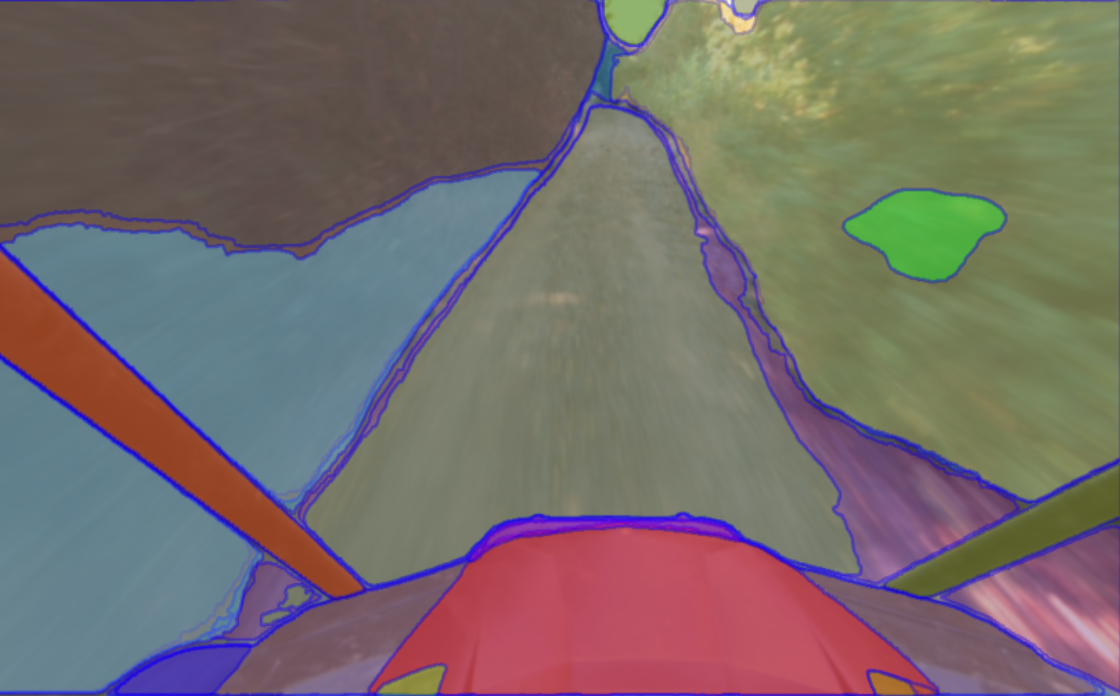}
    \label{subfig:masks}
  \end{subfigure}
   \begin{subfigure}[b]{0.325\linewidth}
    \centering
    \includegraphics[width=\textwidth]{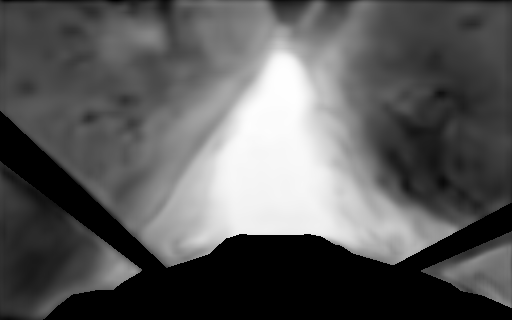}
    \label{subfig:cost}
  \end{subfigure}
  \vspace{-0.8cm}
  \caption{(Top) The Polaris RZR off-road vehicle used for our data collection. (Bottom-left) The vehicle is equipped with multiple RGB-D cameras. (Bottom-middle) Segmentation masks from SAM can disambiguate terrain features. (Bottom-right) Self-supervised learning with mask proposals can achieve robust, fine-grained traversability prediction.}
  \vspace{-0.8cm}
\label{fig:main}
\end{figure}

While various sensors (\textit{e.g.,} LiDAR, IMU, etc.) can be used to capture terrain features, RGB cameras offer dense, high-resolution semantic and geometric information. In addition, learning traversability using RGB cameras offers a unique advantage over other sensor modalities: it allows us to leverage large general-purpose models, trained on massive datasets, to derive robust and expressive features from RGB input.
Such ``Vision Foundation Models" are typically trained on other vision-based tasks, such as semantic segmentation or image classification, but are able to learn rich, mid-level representations that can generalize to other, novel tasks with impressive results~\cite{sam, dino, dinov2}. In particular, the Segment Anything Model (SAM)~\cite{sam}, has demonstrated unparalleled zero-shot performance on instance-based semantic segmentation tasks. Trained on a dataset of 11 million images (with over 1.1 billion mask instances), the model is comprised of a Vision-Transformer (ViT) backbone~\cite{vit}, allowing it to retain fine-grained visual information and generate multiple mask proposals for most objects in the scene (\cref{fig:main}). Such models have the potential to drastically improve the generalization performance of traversability learning for off-road autonomy. To date, however, this has yet to be demonstrated. How to best utilize these models remains an open question.

In this paper, we posit that leveraging mask segments for self-supervision offers a simple yet pragmatic method for bootstrapping traversability learning: assuming that image pixels corresponding to the same object or terrain patch should have a similar level of traversability, class-agnostic semantic masks can provide strong priors for self-supervised learning.
Specifically, we propose a novel method for pixel-level, contrastive self-supervised learning using projected vehicle trajectories in pixel space along with mask proposals from SAM, overcoming the limitation of solely using trajectory masks as shown in~\cref{fig:occlusion_handling}.

We demonstrate the effectiveness of our method on newly collected off-road datasets as our benchmark, on which our method drastically outperforms state-of-the-art baseline methods. Furthermore, while many existing methods validate their approach on off-road but on-trail sequences, ours effectively predicts traversability for both on-/off-trail cases in a number of varied, diverse environments. Lastly, we show how our method can be used for zero- and few-shot traversability learning in new environments not covered in the training data.

%% file: sections/related_work.tex
\section{Related Work}

\vspace{-0.1cm}
\subsection{Self-supervised Traversability Learning}
Given the aforementioned limitations in labeling traversability for off-road terrains, numerous approaches to learning traversability in a self-supervising fashion have been proposed for off-road autonomy~\cite{Zurn2019SelfSupervisedVT, dahlkamp2006self, karnan2023self, castro2023does, daejeon2023, scherer2023, jpl2022}.
Such efforts have made use of a variety of sensory modalities to provide a self-supervised learning signal.
For instance, Castro \textit{et al.}~\cite{castro2023does} used IMU $z$-axis measurements as a traversability score, whereas Seo \textit{et al.}~\cite{seo2023scate} used a combination of proprioceptive sensors and LiDAR to extract vehicle trajectory traces.

Recently, several approaches~\cite{Zurn2019SelfSupervisedVT, scherer2023, jpl2022, daejeon2023, karnan2023self} have proposed to predict traversability from visual information.
Schmid \textit{et al.}~\cite{jpl2022} proposed to use autoencoder-based anomaly detection to classify visual terrain features that the system has not traversed.
Inspired by~\cite{wellhausen2019should}, they aggregate the vehicle footprints and project them into image space to crop out traversed regions.
Then, the autoencoder~\cite{vae} is trained to reconstruct the traversed regions only.
This forces the model to fail on not-traversed areas since they are not observed during the training (\textit{i.e.,} out-of-distribution areas).
Thus, at test time, they translate the reconstruction error into the traversability score.
However, as the authors remarked, this approach can be susceptible to illumination changes and may produce visual artifacts due to the nature of the reconstruction loss.

\begin{figure}[t!]
  \centering
  \includegraphics[width=0.85\linewidth]{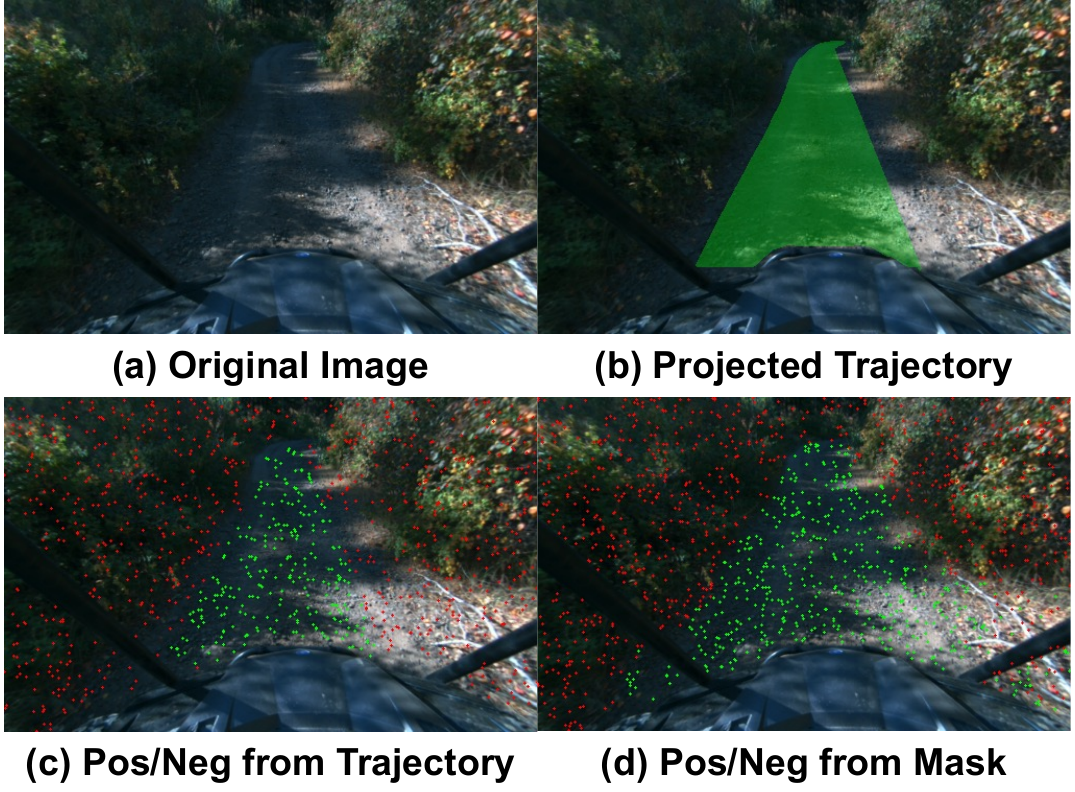}
  \vspace{-0.2cm}
  \caption{Illustration of our occlusion handling and trajectory-/mask-based sampling examples.
  From a given image and trajectory, we use stereo depth to filter out occluded poses and project them into the image space (b). Afterward, we generate random positive samples (bright-green pixels in (c)) and negative samples (red pixels in (c)) within and outside the projected trajectory.
  Using positive samples from trajectory as query points, we obtain a mask prediction from SAM to cover the whole traversable region. Then, we randomly sample positive and negative points using the mask (d).}
  \vspace{-0.8cm}
  \label{fig:occlusion_handling}
\end{figure}
\begin{figure*}[ht!]
  \centering
  \includegraphics[width=0.8\linewidth]{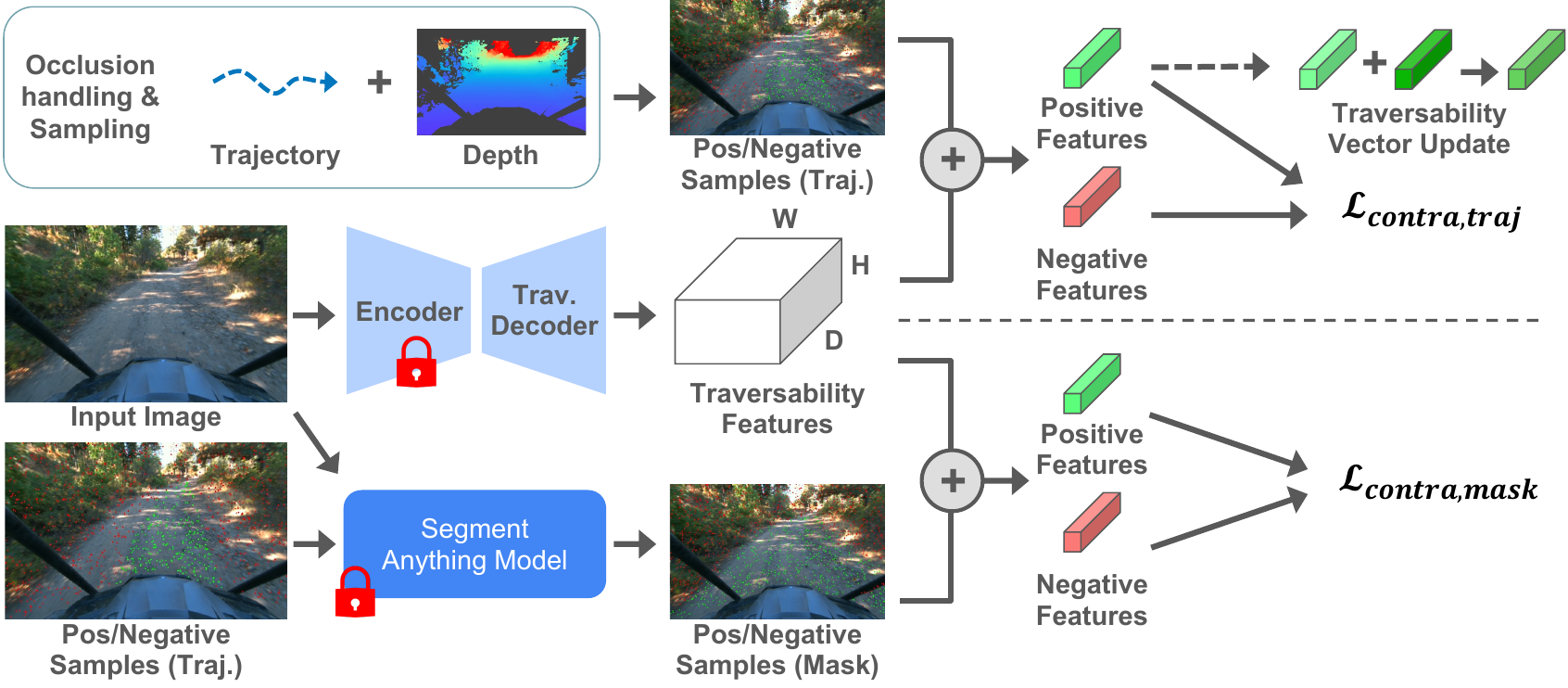}
  \vspace{-0.2cm}
  \caption{Overview of our method. We first incorporate stereo-depth information to filter out occluded trajectory points and then project the trajectory into image space.
  Then, positive and negative points are sampled based on the trajectory and SAM-predicted mask information. 
  We apply a pre-trained image encoder along with the traversability decoder that outputs traversability features.
  Afterward, we extract positive and negative features and apply the trajectory-/mask-based contrastive losses to train the decoder.
  Additionally, we update our traversability vector using a running-average over positive features.
  This updated vector is used to calculate the similarity at test time, which will be directly translated into traversability costs.
  Note that a dashed gray arrow denotes gradient stop, and we do not update our encoder during training.}
  \vspace{-0.8cm}
  \label{fig:method_overview}
\end{figure*}

\vspace{-0.2cm}
\subsection{Contrastive Traversability Learning}
Several approaches have adopted contrastive learning in their traversability learning~\cite{xue2023contrastive, daejeon2023, Zurn2019SelfSupervisedVT, wellhausen2019should}. 
In Xue \textit{et al.}~\cite{xue2023contrastive}, prototype vectors are learned from embedded positive and unlabeled terrain patches based on LiDAR features, which also serve to generate pseudo-labels for an additional supervised classification task.
Additionally, other approaches propose to train a network to generate discriminative feature embeddings either with acoustic features~\cite{Zurn2019SelfSupervisedVT} or proprioceptive sensor readings~\cite{wellhausen2019should} coupled with weakly supervised labels learning to estimate traversability with relatively less manual labeling.
In comparison, our method only utilizes visual input from RGB cameras and does not require other sensor measurements.

Perhaps closest in spirit to our work is Seo \textit{et al.}~\cite{daejeon2023}, where the authors propose to use Positive-Unlabeled (PU) learning~\cite{pulearning} with image-level contrastive learning~\cite{simclr}.
The authors adopt a normalizing-flow~\cite{norm_flow} model and apply a PU learning algorithm for binary classification.
Additionally, the contrastive loss~\cite{simclr} is applied to augmented images to encourage the model to have good image representations.
However, such a contrastive loss may not be sufficient to provide meaningful information in distinguishing traversable and non-traversable areas.
On the contrary, we utilize contrastive learning to promote a model to separate traversable and non-traversable features in the representation space by sampling positive and negative points within and outside the vehicle trajectories.

\vspace{-0.1cm}
\subsection{Vision Foundation Models}
Recent vision foundation models~\cite{dino, dinov2, sam} have shown remarkable performance in both accuracy and generalization.
In particular, SAM~\cite{sam} has demonstrated impressive performance in identifying different object instances in high-resolution images, even in unseen environments.
As many approaches~\cite{trackanything, trackanything2, inpaintanything} have been proposed using pre-trained SAM networks, we also find strong advantages of using SAM.
First, we take advantage of strong generalization performance and well-trained representation space by adopting the SAM image encoder as our backbone.
In addition, we demonstrate the importance of using SAM-predicted mask proposals to improve trajectory-based self-supervised learning.
Since vehicle trajectories cannot cover all traversable areas appearing in the image, these masks provide auxiliary self-supervised labels outside traversed regions.
As shown in~\cref{fig:occlusion_handling} (c) and (d), incorporating SAM-predicted masks allows us to cover missing traversable areas where manual labeling efforts would previously be required.

%% file: sections/method.tex
\section{Method}
In this section, we first describe trajectory projection and occlusion handling for positive-label generation, followed by trajectory-level and mask-level contrastive loss definitions leveraging SAM mask predictions.
We then elaborate on traversability vector estimation and conversion to a traversability metric.
The overview of our method is illustrated in~\cref{fig:method_overview}.
 
\begin{figure*}[ht!]
  \centering
  \includegraphics[width=0.82\linewidth]{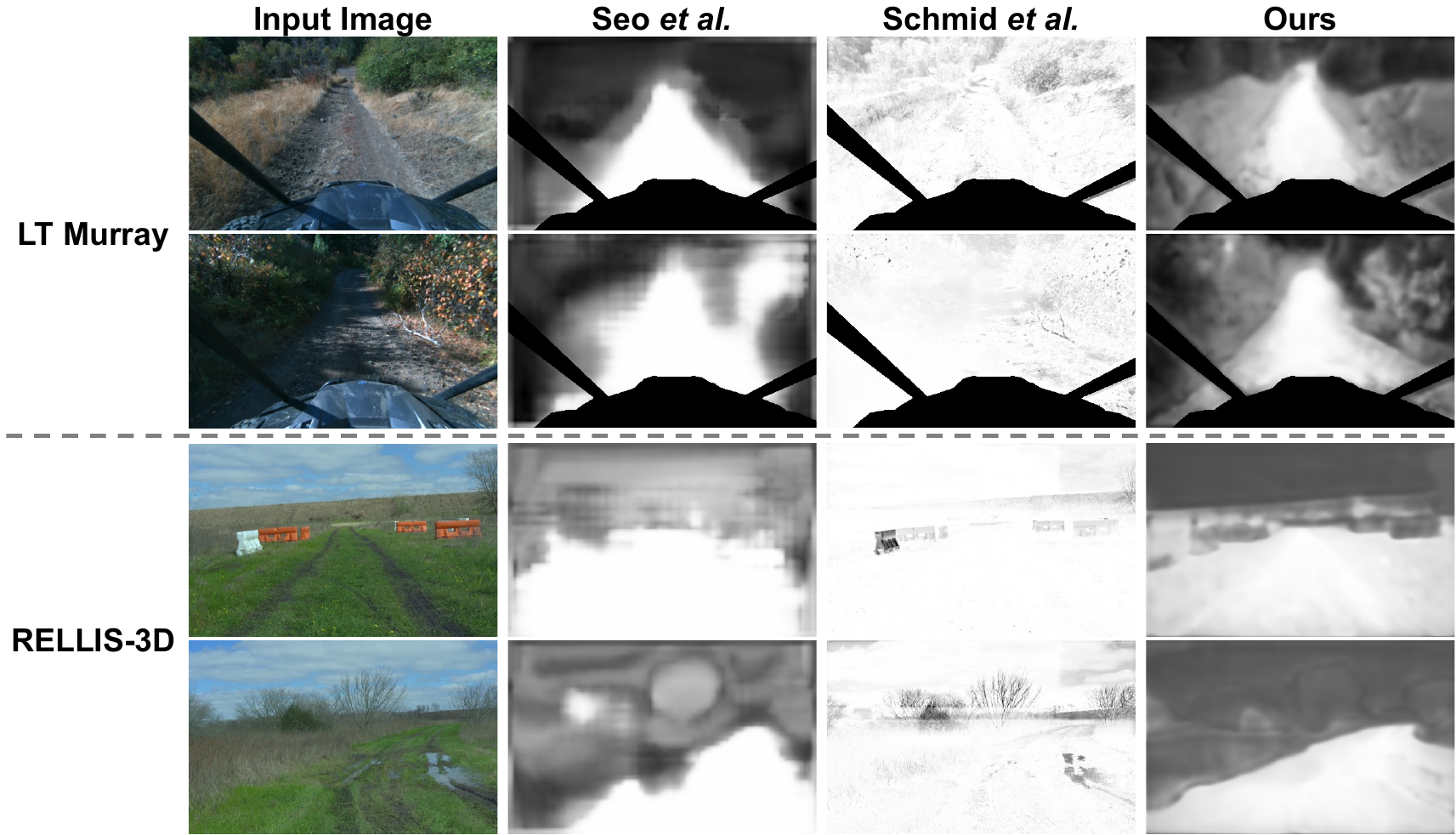}
  \vspace{-0.1cm}
  \caption{Qualitative results of Seo \textit{et al.}~\cite{daejeon2023}, Schmid \textit{et al.}~\cite{jpl2022}, and our method on RELLIS-3D and LT Murray datasets. We strongly encourage readers to view the supplementary video for more detailed qualitative results.}
  \label{fig:qualitative_trail}
  \vspace{-0.8cm}
\end{figure*}
  
\subsection{Trajectory Projection and Occlusion Handling}
We use vehicle trajectories as a self-supervision signal for learning traversability, as shown in~\cref{fig:method_overview}.
Specifically, with given poses of left/right wheels $P_{t:t+T}$ in the global frame from time $t$ to $t+T$, we project the trajectory into the image space by
\begin{equation}
    P_{t:t+T}^I = K[R|t]P_{t:t+T},
\end{equation}
where $K$ denotes an intrinsic matrix of the camera, $R, t$ indicate rotation and translation of an extrinsic matrix, and $T$ denotes the time horizon.
We filter out occluded points by using a stereo depth estimate $D_t$ at time $t$:
\begin{equation}
    P^{I, \text{filtered}}_{t:t+T} = \{p | p\in P^I_{t:t+T},\, p_z \leq D_t(p) \}.
\end{equation}
Note that the depth estimate $D_t$ is obtained from a stereo depth sensor, aligned with the RGB camera.
We then calculate the contour of the projected left/right trajectories and fill in the contour to complete the trajectory region (\cref{fig:occlusion_handling} (b)).
Finally, we randomly sample positive points $\mathbb{P}_\text{traj}$ within the completed trajectory and negative points $\mathbb{N}_\text{traj}$ outside the trajectory (\cref{fig:occlusion_handling} (c)).

However, as shown in Fig.~\ref{fig:occlusion_handling} (c), the projected trajectory often covers only a portion of the traversable terrain.
Such a gap between the projected trajectory and the actual traversable region adversely affects the training since negative points can be sampled from the gap.
To mitigate this, we use mask predictions from SAM as an additional signal for traversability. 
Specifically, we obtain the mask predictions from SAM~\cite{sam} by using positive samples $\mathbb{P}_\text{traj}$ from the previous step as query points.
We select a mask from the proposals if its area is larger than a certain threshold, and it has the highest confidence.
Then, analogous to the previous sampling step, we sample positive (\textit{i.e.,} $\mathbb{P}_\text{mask}$) and negative (\textit{i.e.,} $\mathbb{N}_\text{mask}$) points within and outside the predicted mask.
As shown in~\cref{fig:occlusion_handling} (d), positive samples from the predicted mask successfully cover the whole traversable region.

\subsection{Contrastive Learning for Traversability}
We define a traversability prediction model $f(\mathbf{x}) = h \circ g(\mathbf{x})$ that predicts traversability features $\mathbf{F}\in\mathbb{R}^{H\times W \times D}$ from a given image $\mathbf{x}\in\mathbb{R}^{H\times W \times 3}$, where $D$ denotes a dimension of traversability features.
The model $f(\mathbf{x})$ is composed of a pre-trained image encoder $g(\cdot)$ and a traversability decoder $h(\cdot)$.
We adopt the pre-trained image encoder from SAM~\cite{sam}, leveraging its generalized latent feature representations, and we do not update the encoder during training.

With the obtained positive and negative samples from the previous step, we apply contrastive losses to train the traversability decoder.
Specifically, with traversability features $\mathbf{F}$, a set of positive samples $\mathbb{P}$, and a set of negative samples $\mathbb{N}$, the contrastive loss is defined as:
\begin{align}
    &\mathcal{L}_\text{contra}(\mathbf{F}, \mathbb{P}, \mathbb{N}) =\\ \nonumber
    &-\frac{1}{N (N - 1)} \sum_{s_i\in\mathbb{P}}\sum_{s_j\in\mathbb{P}}\mathds{1}(i \neq j)\log\frac{\exp{(\mathbf{F}_{s_i}^\intercal\cdot \mathbf{F}_{s_j} / \tau)}}{\sum_{s_k\in\mathbb{N}}\exp{(\mathbf{F}_{s_i}^\intercal\cdot \mathbf{F}_{s_k} / \tau)}},
\end{align}
with pixel-level features $\mathbf{F}_{s}\in \mathbb{R}^{D\times 1}$ at pixel $s$. Here, $N = |\mathbb{P}|$, and $\tau$ denotes a temperature scalar.
We normalize the traversability features $\mathbf{F}$ along with the $D$ dimension before applying the loss.

Finally, our final loss is represented as
\begin{align}
    \mathcal{L} =& (1-\omega_\text{mask})\ \mathcal{L}_\text{contra}(\mathbf{F}, \mathbb{P}_\text{traj}, \mathbb{N}_\text{traj})\\ \nonumber
    &+ \omega_\text{mask}\ \mathcal{L}_\text{contra}(\mathbf{F}, \mathbb{P}_\text{mask}, \mathbb{N}_\text{mask}),
\end{align}
with weight $\omega_{mask} \in \left[0, 1\right]$.

\subsection{Computing Traversability Cost}
While training the model with contrastive losses, we estimate a traversability vector $\mathbf{z}\in\mathbb{R}^D$, for converting traversability features to cost at inference time.
Specifically, we update the traversability vector $\mathbf{z}$ using exponential moving averaging (EMA),
\begin{equation}
    \mathbf{z} = \alpha \mathbf{z} + (1 - \alpha)\frac{1}{|\mathbb{P}_\text{traj}|}\sum_{s\in\mathbb{P}_\text{traj}}\mathbf{F}_{s},
\end{equation}
where $\alpha\in[0, 1]$ denotes a momentum value.
The vector $\mathbf{z}$ and feature $\mathbf{F}_s$ are normalized before and after the EMA operation.
Note that we maintain one traversability vector. This is because the contrastive losses will optimize the traversability decoder to map the SAM feature vectors of traversable regions to have the same representation while distinguishing them from negative feature vectors.

At test time, we convert the traversability features $\mathbf{F}$ to costs $\mathbf{C}\in\mathbb{R}^{H\times W}$ by calculating cosine similarity between $\mathbf{F}$ and $\mathbf{z}$ (\textit{i.e.,} $\mathbf{C} = \mathbf{Fz}$).
Since cosine similarity has a range of $[-1, 1]$, the predicted cost $\mathbf{C}$ is always bounded.

%% file: sections/experiments.tex
\begin{figure*}[ht!]
  \centering
  \includegraphics[width=0.82\linewidth]{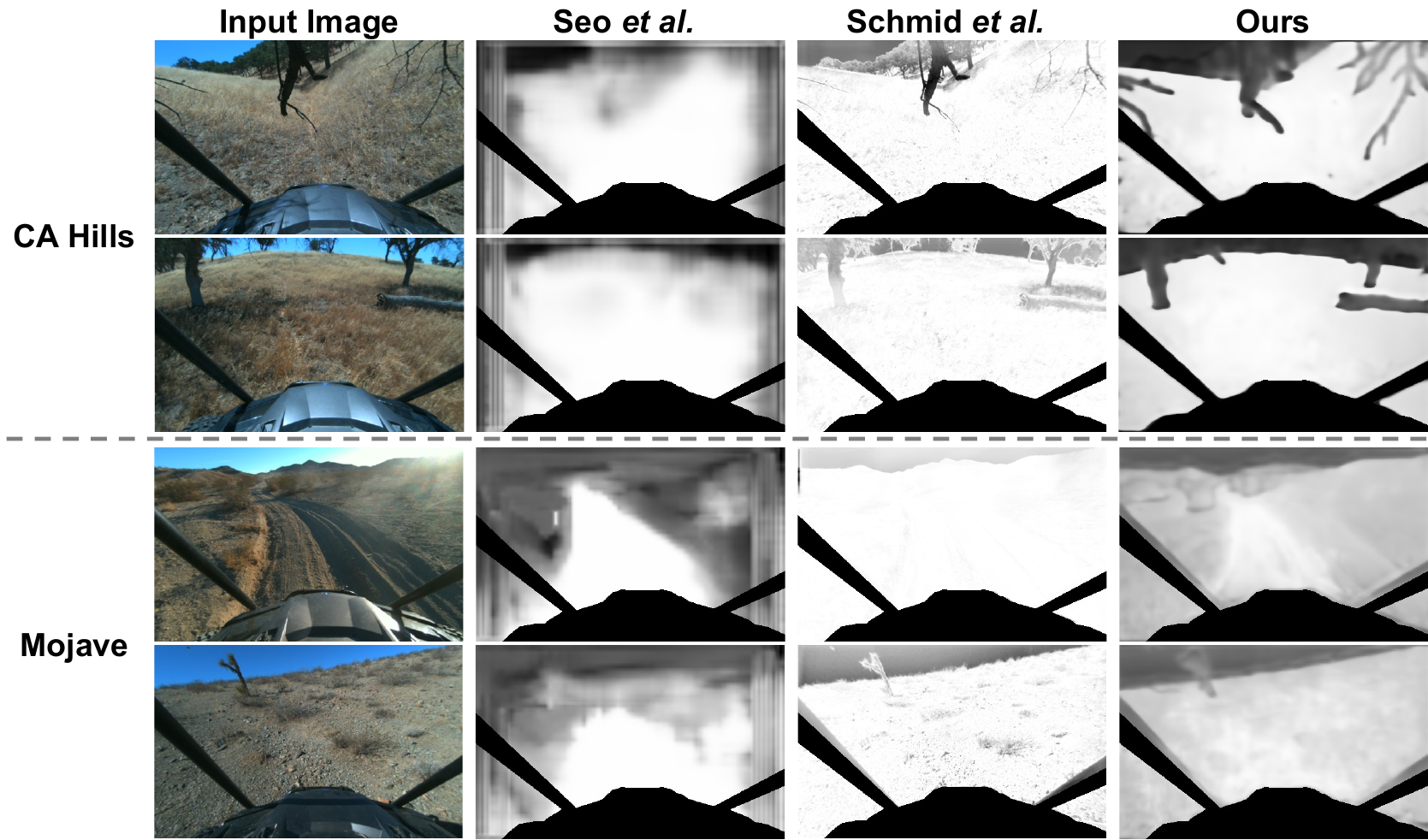}
  \vspace{-0.15cm}
  \caption{Qualitative results of Seo \textit{et al.}~\cite{daejeon2023}, Schmid \textit{et al.}~\cite{jpl2022}, and our method on CA Hills and Mojave Desert validation sequences. We strongly encourage readers to view the supplementary video for more detailed qualitative results.
}
  \vspace{-0.8cm}
  \label{fig:qualitative_main}
\end{figure*}

\section{Experiments}
\label{sec:experiments}
\subsection{Dataset Collection and Processing}
Our datasets cover both on-trail and off-trail environments with varied types of terrain.
Data collection was performed at the following sites in the US: LT Murray (WA), Mojave Desert (CA), and California Hills (CA).

\noindent\textbf{LT Murray (WA)} sequence is composed of on-trail scenes, where the trails go through dense and sparse vegetation with bushes and trees.
We recorded 40 miles of on-trail data.

\noindent\textbf{Mojave Desert (CA)} consists of both on-trail and off-trail scenes with rocks, small bushes, Joshua trees, and cacti.
We recorded two different runs by following the trail or driving through small bushes and rocks.

\noindent\textbf{CA Hills} is composed of off-trail scenes on grassy hills populated with trees.
We recorded two different runs by driving the terrain.

We collected images, stereo depths, LiDAR point clouds, and vehicle poses from each run and ran an offline SLAM algorithm~\cite{cartographer} to obtain precise state estimations.
For evaluation purposes, we labeled LiDAR point clouds and generated segmentation ground truth in BEV.
We also set a manual cost per class in BEV for hyper-parameter selection by running real-world experiments.

\vspace{-0.1cm}
\subsection{Baselines}
For fair evaluations, we compare our method with Seo \textit{et al.}~\cite{daejeon2023} and Schmid \textit{et al.}~\cite{jpl2022}.
Both methods use vehicle trajectories to self-supervise the model to learn traversability.
Seo \textit{et al.} adopt a normalizing-flow model~\cite{norm_flow} along with an image-level contrastive loss to train traversability, while Schmid \textit{et al.} crop the trajectory regions in images and train variational auto-encoder (VAE)~\cite{vae} to reconstruct the traversed regions.
Afterward, the reconstruction error is translated into traversability scores at test time.

\vspace{-0.1cm}
\subsection{Implementation details}
We adopt the encoder of the ViT-H SAM model as our image encoder and freeze it during training.
Note that our full process, from image to prediction, runs around 6 Hz on a single NVIDIA A100 GPU.
For training, we use a learning rate of 1e-3 and a batch size of 2 and train the model for 1 epoch.
We sample 256/1024 positive/negative points for the trajectory contrastive loss, and sample 512/1024 positive/negative points for the mask contrastive loss.
Positive/negative samples on the ego vehicle are excluded.
We use $\alpha=0.999$ to update the traversability vector and $T=300$ for trajectory projection.
We set the temperature $\tau=0.05$ and the mask contrastive loss weight $\omega_\text{mask}=0.05$.
These hyper-parameters (\textit{i.e.,} $\tau$ and $\omega_\text{mask}$) have been selected based on the averaged L1 error between predicted costs and manual costs obtained from ground-truth segmentation labels.
Note that we project the predicted costs into BEV using stereo depths for hyper-parameter selection since the ground truth labels are in BEV.

\vspace{-0.1cm}
\subsection{Qualitative Results}
\cref{fig:qualitative_trail} shows predictions of ours and baseline methods on the RELLIS-3D and LT Murray, which are on-trail datasets.
While Seo \textit{et al.}~\cite{daejeon2023} successfully predicts trajectories on trails, it fails to predict lethal objects clearly and distinguish the costs between different semantics such as trees, bushes, and traffic barriers.
Schmid \textit{et al.}~\cite{jpl2022} predicts trails as low costs but fails to mark surrounding lethal objects as high costs.
On the other hand, our method clearly marks the trails as low costs and maintains subtle differences well in objects of off-trail terrain.

The strength of our method becomes clearer when it is applied to off-trail scenarios.
\cref{fig:qualitative_main} illustrates cost prediction results on CA Hills and Mojave Desert sequences, which contain off-trail images.
Baselines struggle to find traversable regions correctly or to predict the lethal object as a high cost.
Our approach not only distinguishes traversable regions from lethal objects correctly but also assigns different costs for different semantic objects.
We conjecture this is because each semantic class has a different frequency to be sampled as positive.
For example, the grass can be sampled as positive if the vehicle goes over it, but the tree cannot be sampled because the vehicle cannot pass through it.

\begin{table}[ht!]
\begin{center}
\footnotesize
\vspace{-0.1cm}
\scalebox{0.85}{
\begin{tabular}{c|c|c|c|c|c|c|c}
\toprule
Methods & AUROC$\uparrow$ & AP$\uparrow$ & F1$\uparrow$ & FPR$\downarrow$ & FNR$\downarrow$ & Pre.$\uparrow$ & Rec.$\uparrow$ \\
\drule
Schmid \textit{et al.} & 0.713 & 0.666 & 0.601 & 0.455 & 0.273 & 0.513 & 0.727 \\
Seo \textit{et al.} & 0.931 & 0.947 & 0.888 & 0.262 & \textbf{0.063} & 0.844 & \textbf{0.937}\\
\textbf{Ours} & \textbf{0.959} & \textbf{0.975} & \textbf{0.929} & \textbf{0.100} & 0.075 & \textbf{0.934} & 0.925\\
\bottomrule
\end{tabular}}
\vspace{-0.1cm}
\caption{Evaluation results on the RELLIS-3D dataset~\cite{rellis}. Pre. and Rec. stand for precision and recall.}
\vspace{-0.8cm}
\label{tab:rellish}
\end{center}
\end{table}

\begin{table}[ht!]
\begin{center}
\vspace{-0.1cm}
\footnotesize
\scalebox{0.85}{
\begin{tabular}{c|c|c|c}
\toprule
\multirow{2}{*}{Methods} & \multicolumn{3}{c}{Collision Rate $\downarrow$} \\
\cmidrule{2-4}
& LT Murray & CA Hills & Mojave Desert \\
\drule
Schmid \textit{et al.} & 0.161 & 0.106 &  0.001 \\
Seo \textit{et al.} & 0.128 & 0.110 & \textbf{0.000}\\
\textbf{Ours} & \textbf{0.107} & \textbf{0.056} & 0.001\\
\bottomrule
\end{tabular}}
\vspace{-0.1cm}
\caption{Collision rates of different methods on the LT Murray, CA Hills, and Mojave Desert datasets.}
\vspace{-0.8cm}
\label{tab:MPPI_ours}
\end{center}
\end{table}

\begin{figure}[ht!]
  \centering
  \vspace{-0.2cm}
  \includegraphics[width=\linewidth]{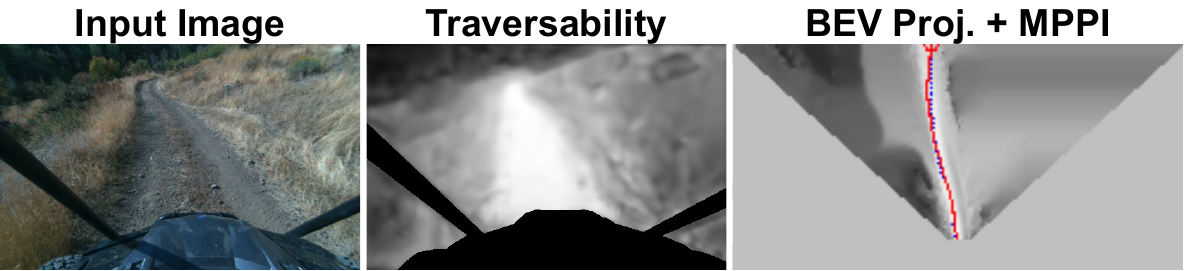}
  \vspace{-0.6cm}
  \caption{Example of the projected costs and MPPI results. The blue trajectory is the expert demonstration, and the red trajectory is a result of successive application of MPPI-optimized control inputs.}
  \label{fig:qualitative_MPPI}
  \vspace{-0.5cm}
\end{figure}

\subsection{Quantitative Results}
\noindent\textbf{RELLIS-3D dataset}
We evaluate our method and baselines on the RELLIS-3D dataset~\cite{rellis}, which is a publicly available labeled dataset.
We adopt AUROC (area under receiver operating characteristic), AUPRC (area under precision-recall curve), F1 score, FPR (false positive rate), FNR (false negative rate), Precision, and Recall metrics for evaluation.
Note that we report FPR, FNR, Precision, and Recall metrics at the best threshold that achieves the highest F1 score.

\cref{tab:rellish} shows the comparisons on the RELLIS-3D dataset. Our method outperforms baselines by a large margin overall, except for FNR and Recall.
However, even in such cases, the gaps are marginal (\textit{i.e.,} 0.012 in both FNR and Recall) compared to our performance gains in other metrics.

\noindent\textbf{Our datasets}
Since our datasets do not have labels in image space but in BEV, we project the predicted costs into BEV and inpaint the missing costs using the nearest values.
Then, we run a model predictive control (MPC) algorithm to obtain the optimal trajectory based on the predicted costmaps (\cref{fig:qualitative_MPPI}), over all the validation sequences.
Specifically, we adopt model predictive path integral (MPPI)~\cite{mppi} as our MPC algorithm and measure the number of collisions with lethal objects over the number of successful MPPI runs (\textit{i.e.,} Collision Rate).
To check collision, we use the ground truth labels in BEV and follow the optimal trajectory obtained from running MPPI on predicted costmaps.
To achieve a low collision rate, it is important to find all the lethal objects appearing in the image while finding traversable areas correctly.
As reported in~\cref{tab:MPPI_ours}, our method outperforms other baselines in LT Murray and CA Hills datasets by a large margin.
In the case of the Mojave Desert dataset, all three methods show a similar performance achieving nearly 0 collision rates.

\begin{figure}[h!]
  \centering
  \vspace{-0.3cm}
  \includegraphics[width=\linewidth]{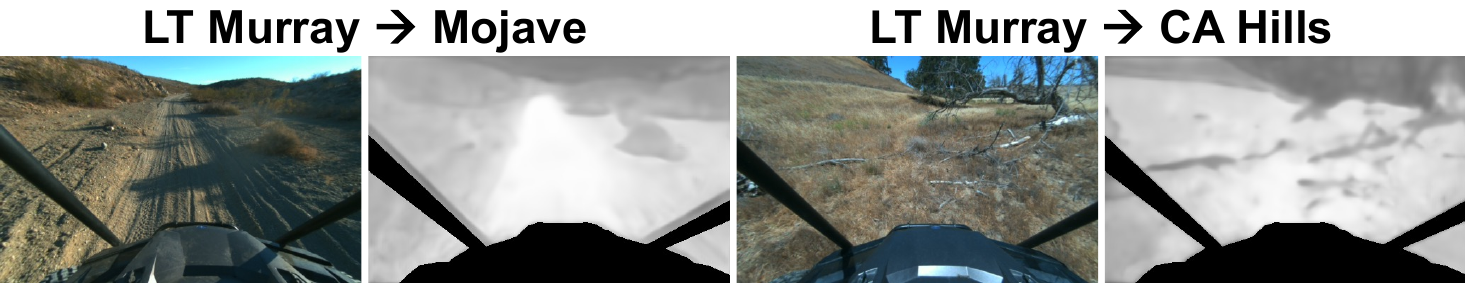}
  \caption{Zero-shot generalization results of our method in LTMurray to Mojave Desert (left) and LTMurray to CA Hills (right) scenarios.}
  \label{fig:qualitative_zeroshot}
  \vspace{-0.5cm}
\end{figure}

\begin{figure}[h!]
  \centering
  \vspace{-0.3cm}
  \includegraphics[width=\linewidth]{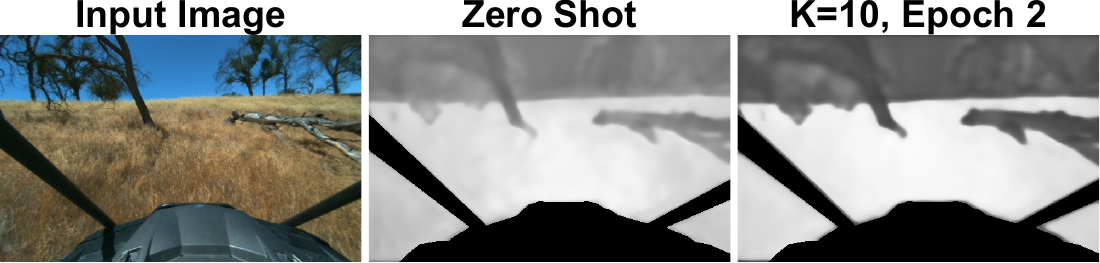}
  \caption{Few-shot adaptation results of our method in the LTMurray to CA Hills scenario. 
  Iterating two epochs over 10 samples significantly improves the traversability predictions.}
  \label{fig:qualitative_fewshot}
  \vspace{-0.8cm}
\end{figure}

\begin{table}[ht!]
\begin{center}
\footnotesize
\scalebox{0.85}{
\begin{tabular}{c|c|c|c}
\toprule
\multirow{2}{*}{Datasets} & \multicolumn{3}{c}{Collision Rate $\downarrow$} \\
\cmidrule{2-4}
& In-domain & Zero-shot & Few-shot \\
\drule
Mojave Desert & \textbf{0.001} & \textbf{0.001} & - \\
CA Hills & 0.056 & 0.092 & \textbf{0.048}\\
\bottomrule
\end{tabular}}
\vspace{-0.1cm}
\caption{Evaluation results on the LT Murray, CA Hills, and Mojave Desert datasets.}
\vspace{-0.7cm}
\label{tab:zeroshot}
\end{center}
\end{table}

\vspace{-0.1cm}
\subsection{Generalization to New Environments}
A strong advantage of having well-trained visual representations is generalization to out-of-distribution environments.
To demonstrate the robustness of our method in new environments, we train a model on the on-trail LT Murray dataset and test it on the Mojave Desert and CA Hills.
As illustrated in~\cref{fig:qualitative_zeroshot}, our method generalizes surprisingly well without any adaptation.
The model successfully finds a cactus and a small tree in Mojave and identifies logs lying on the ground in CA Hills, while predicting low cost on traversable areas.
The MPPI evaluation results reported in~\cref{tab:zeroshot} also align with such observations.
The few-shot adaptation result in CA Hills outperforms the in-domain results. This demonstrates that our method can effectively transfer from one domain to another, unseen domain.

\begin{table}[ht!]
\begin{center}
\footnotesize
\scalebox{0.85}{
\begin{tabular}{c|c|c|c|c|c|c|c}
\toprule
Methods & AUROC$\uparrow$ & AP$\uparrow$ & F1$\uparrow$ & FPR$\downarrow$ & FNR$\downarrow$ & Pre.$\uparrow$ & Rec.$\uparrow$ \\
\drule
w/o mask  & 0.920 & 0.945 & 0.892 & 0.240 & 0.068 & 0.855 & 0.932\\
\textbf{w/ mask} & \textbf{0.959} & \textbf{0.975} & \textbf{0.929} & \textbf{0.100} & 0.075 & \textbf{0.934} & 0.925\\
Solely w/ mask & 0.955 & 0.972 & 0.927 & 0.129 & \textbf{0.063} & 0.917 & \textbf{0.937}\\
\bottomrule
\end{tabular}}
\vspace{-0.1cm}
\caption{Effect of applying the mask-based loss to our method in the RELLIS-3D dataset. Incorporating mask information significantly improves the overall performance.}
\vspace{-0.6cm}
\label{tab:effect_of_mask}
\end{center}
\end{table}

\begin{figure}[ht!]
  \centering
  \includegraphics[width=0.86\linewidth]{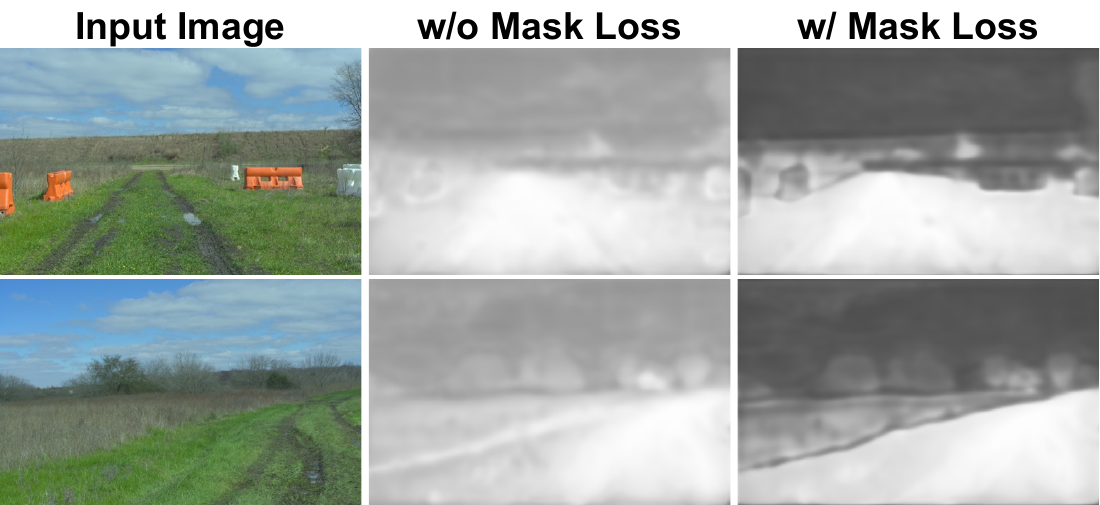}
  \vspace{-0.1cm}
  \caption{Qualitative results on the RELLIS-3D dataset with and without the mask loss. Incorporating mask information significantly improves discriminability between traversable and non-traversable areas.}
  \label{fig:qualitative_mask_ablation}
  \vspace{-0.6cm}
\end{figure}

\subsection{Ablation Study}
\label{subsec:ablation}
\noindent\textbf{Effectiveness of using masks}
Using vehicle trajectories alone for self-supervision is insufficient since the trajectory cannot cover the whole traversable region.
This ablation study aims to reveal how mask-based self-supervision is effective in addressing such a concern by comparing results with and without mask-based loss.
As reported in \cref{tab:effect_of_mask}, applying the mask-based loss brings significant improvements in overall metrics.
Additionally, as shown in~\cref{fig:qualitative_mask_ablation}, the model without mask information overfits vehicle trajectories and marks left and right side regions as high costs even though they are indeed traversable.
On the other hand, once we apply the mask-based loss, the model correctly identifies traversable regions and marks them as low costs, mitigating the overfitting problem.
Solely using mask-based loss also presents improvements over the one that did not use any mask at all, but it is still behind our final version in most of the metrics.
We also observe that false mask proposals from SAM adversely affect the training even though the number of them is very few.

%% file: sections/conclusion.tex
\section{Conclusion}
In this paper, we proposed a novel self-supervised approach to learning traversability in image space using contrastive learning. We show that with the addition of mask-based regularization, guided by robust segmentation proposals, generalization of traversability predictions can be drastically improved and deliver state-of-the-art performance. This is further emphasized by results in out-of-distribution environments.
As for future work, we hope to incorporate temporal sequences of image data and learn-able in-painting to improve the quality of predictions, and investigate extending the approach for online adaptation.

%% file: sections/acknowledgement.tex
\section{Acknowledgements}
This research was developed with funding from the Defense Advanced Research Projects Agency (DARPA). The views, opinions and/or findings expressed are those of the author and should not be interpreted as representing the official views or policies of the Department of Defense or the U.S. Government.